\documentclass[11pt,a4paper]{article}
\usepackage[hyperref]{mrp19}
\usepackage{bbm}
\usepackage{times}
\usepackage{latexsym}
\usepackage{graphicx}
\usepackage{amsmath}
\usepackage{amsfonts}
\usepackage{url}
\usepackage{mathtools}
\usepackage{verbatim}
\usepackage[utf8]{inputenc}
\usepackage[shellescape]{gmp}
\usepackage{dot2texi}
\usepackage{tikz}
\usetikzlibrary{shapes,arrows}
\usepackage{natbib}
\usepackage{caption}
\usepackage{subcaption}
\usepackage{tikz-dependency}
\usepackage{wrapfig}
\usepackage{multirow}
\usepackage{pgfplots}
\usepackage{filecontents}
\usepackage[T1]{fontenc}
\usepackage{subfiles}
\usepackage{readarray}
\usepackage{xcolor}
\usepackage{import}
\usepackage{url}
\usepackage{bm}
\aclfinalcopy 

\usepackage[super]{nth}

\title{ShanghaiTech at MRP~2019: Sequence-to-Graph Transduction with Second-Order Edge Inference for Cross-Framework Meaning Representation Parsing}
\author{Xinyu Wang, Yixian Liu, Zixia Jia, Chengyue Jiang, Kewei Tu \\
    School of Information Science and Technology,\\
    ShanghaiTech University, Shanghai, China\\
  \texttt{\{wangxy1,liuyx,jiazx,jiangchy,tukw\}@shanghaitech.edu.cn} \\
  }

\date{}

\begin{document}
\maketitle
\begin{abstract}
  This paper presents the system used in our submission to the \textit{CoNLL 2019 shared task: Cross-Framework Meaning Representation Parsing}. Our system is a graph-based parser which combines an extended pointer-generator network that generates nodes and a second-order mean field variational inference module that predicts edges. Our system achieved \nth{1} and \nth{2} place for the DM and PSD frameworks respectively on the in-framework ranks and achieved \nth{3} place for the DM framework on the cross-framework ranks.
\end{abstract}

\section{Introduction}
The goal of the \textit{Cross-Framework Meaning Representation Parsing} (MRP 2019, \citet{Oep:Abe:Haj:19}) is learning to parse text to multiple formats of meaning representation with a uniform parsing system. The task combines five different frameworks of graph-based meaning representation. DELPH-IN MRS Bi-Lexical Dependencies (DM) \cite{Iva:Oep:Ovr:12} and Prague Semantic Dependencies (PSD) \cite{hajic-etal-2012-announcing,miyao-etal-2014-house} first appeared in SemEval 2014 and 2015 shared task Semantic Dependency Parsing (SDP) \cite{oepen-etal-2014-semeval,oepen-etal-2015-semeval}. 
Elementary Dependency Structures (EDS) \cite{oepen-lonning-2006-discriminant} is the origin of DM Bi-Lexical Dependencies, which encodes English Resource Semantics \cite{flickinger-etal-2016-english} in a variable-free semantic dependency graph. 
Universal Conceptual Cognitive Annotation (UCCA) \cite{abend-rappoport-2013-universal} targets a level of semantic granularity that abstracts away from syntactic paraphrases. 
Abstract Meaning Representation (AMR) \cite{banarescu-etal-2013-abstract} targets to abstract away from syntactic representations, which means that sentences have similar meaning should be assigned the same AMR graph. One of the main differences between these frameworks is their level of abstraction from the sentence. SDP is a bi-lexical dependency graph, where graph nodes correspond to tokens in the sentence. EDS and UCCA are general forms of anchored semantic graphs, in which the nodes are anchored to arbitrary spans of the sentence and the spans can have overlaps. AMR is an unanchored graph, which does not consider the correspondence between nodes and the sentence tokens. The shared task also provides a cross-framework metric which evaluates the similarity of graph components in all frameworks.

Previous work mostly focused on developing parsers that support only one or two frameworks while few work has explored cross-framework semantic parsing. \citet{peng2017deep}, \citet{stanovsky-dagan-2018-semantics} and \citet{kurita-sogaard-2019-multi} proposed methods learning jointly on the three frameworks of SDP and \citet{peng2018learning} further proposed to learn from different corpora. \citet{hershcovich-etal-2018-multitask} converted UCCA, AMR, DM and UD (Universal Dependencies) into a unified DAG format and proposed a transition-based method for UCCA parsing.

In this paper, we present our system for MRP 2019. Our system is a graph-based method which combines an extended pointer-generator network introduced by \citet{zhang-etal-2018-stog} to generate nodes for EDS, UCCA and AMR graphs and a second-order mean field variational inference module introduced by \citet{wang-etal-2019-second} to predict edges for all the frameworks. According to the official results, our system gets 94.88 F1 score in the cross-framework metric for DM, which is the \nth{3} place in the ranking. For in-framework metrics, our system gets 92.98 and 81.61 labeled F1 score for DM and PSD respectively, which are ranked \nth{1} and \nth{2} in the ranking.

\section{Data Processing}
\begin{figure}[t!]
\centering
\begin{subfigure}{\linewidth}
\centering
\includegraphics[width=0.55\linewidth]{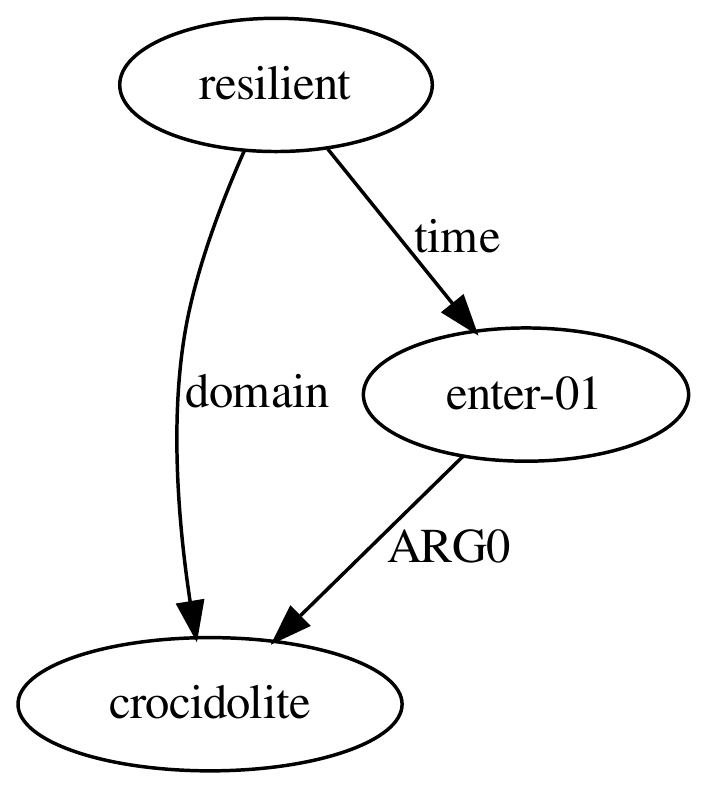}
\caption{Before conversion.}
\end{subfigure}{}
\begin{subfigure}{\linewidth}
\centering
\includegraphics[width=0.75\linewidth]{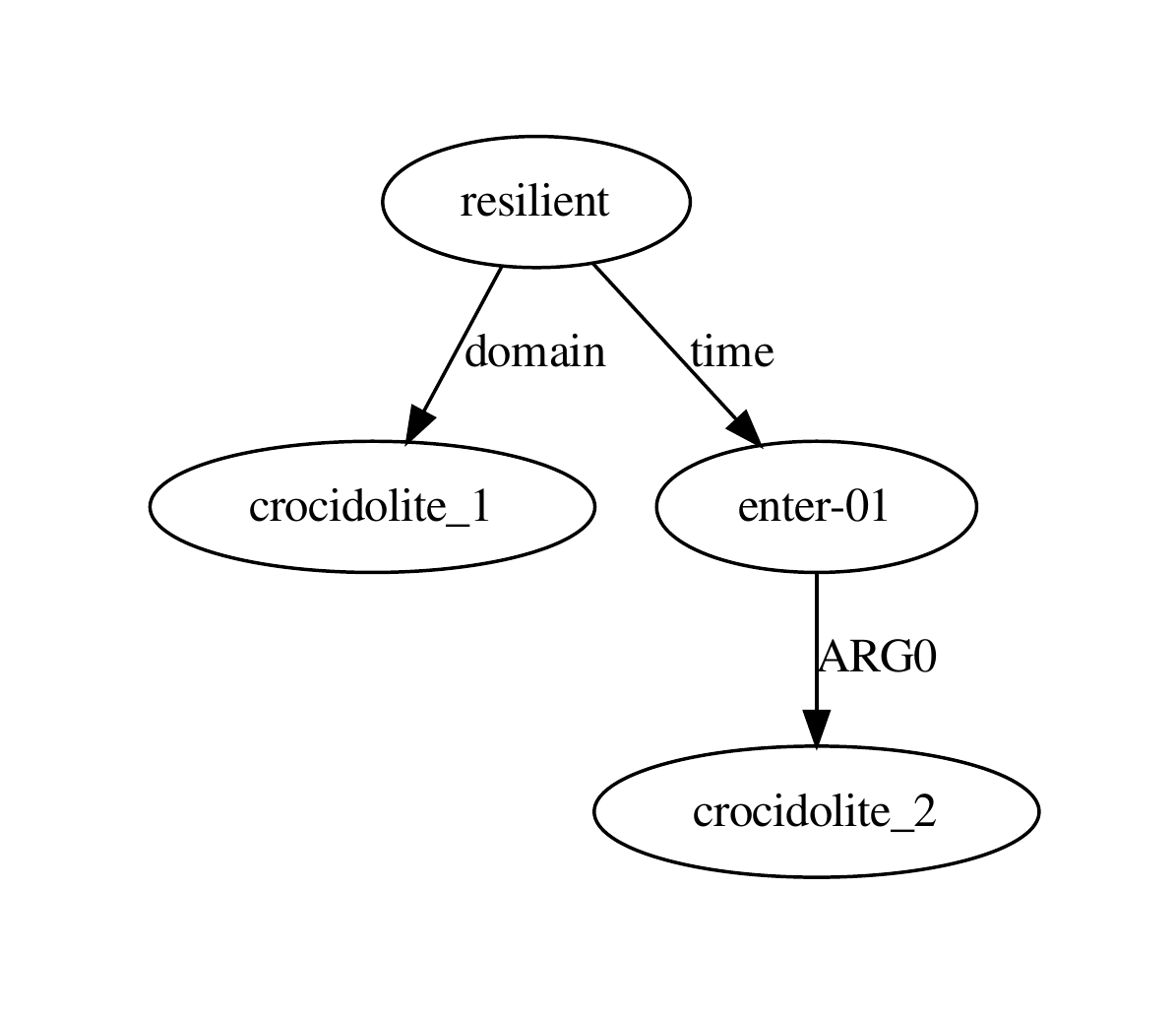}
\caption{After conversion.}
\end{subfigure}{}
\caption{An example of converting AMR graphs into tree structures. This is a sub-graph of sentence \textit{\#20003002}.}
\label{fig:2tree}
\end{figure}
In this section, we introduce our data pre-processing and post-processing in our system for all the frameworks. We use sentence tokenizations, POS tags and lemmas from the official companion data and named entity tags extracted by Illinois Named Entity Tagger \cite{RatinovRo09} in the official `white-list'. We follow \citet{zhang-etal-2018-stog} to convert each EDS, UCCA, and AMR graph to a tree through duplicating the nodes that have multiple edge entrances, An example is shown in Fig. \ref{fig:2tree}. The node sequences for EDS, UCCA and AMR are decided by depth-first search that starts from the root node and sorts neighbouring nodes in alphanumerical order.

\subsection{AMR Data Processing}
Our data processing follows \citet{zhang-etal-2018-stog}. In pre-processing, we remove the senses, wiki links and polarity attributes in AMR nodes, and replace the sub-graphs of special named entities, such as names, places, time, with anonymized words. The corresponding phrases in the sentences are also anonymized. A mapping from NER tags to these entities is built to process the test data. 

In post-processing, we generate the AMR sub-graphs from the anonymized words. Then we assign the senses, wiki links and polarity attributes with the method in \citet{zhang-etal-2018-stog}.

\subsection{EDS and UCCA Data Processing}
\label{sec:2.2}
\begin{figure}[t!]
\centering
\begin{subfigure}{\linewidth}
\centering
\includegraphics[width=0.99\linewidth]{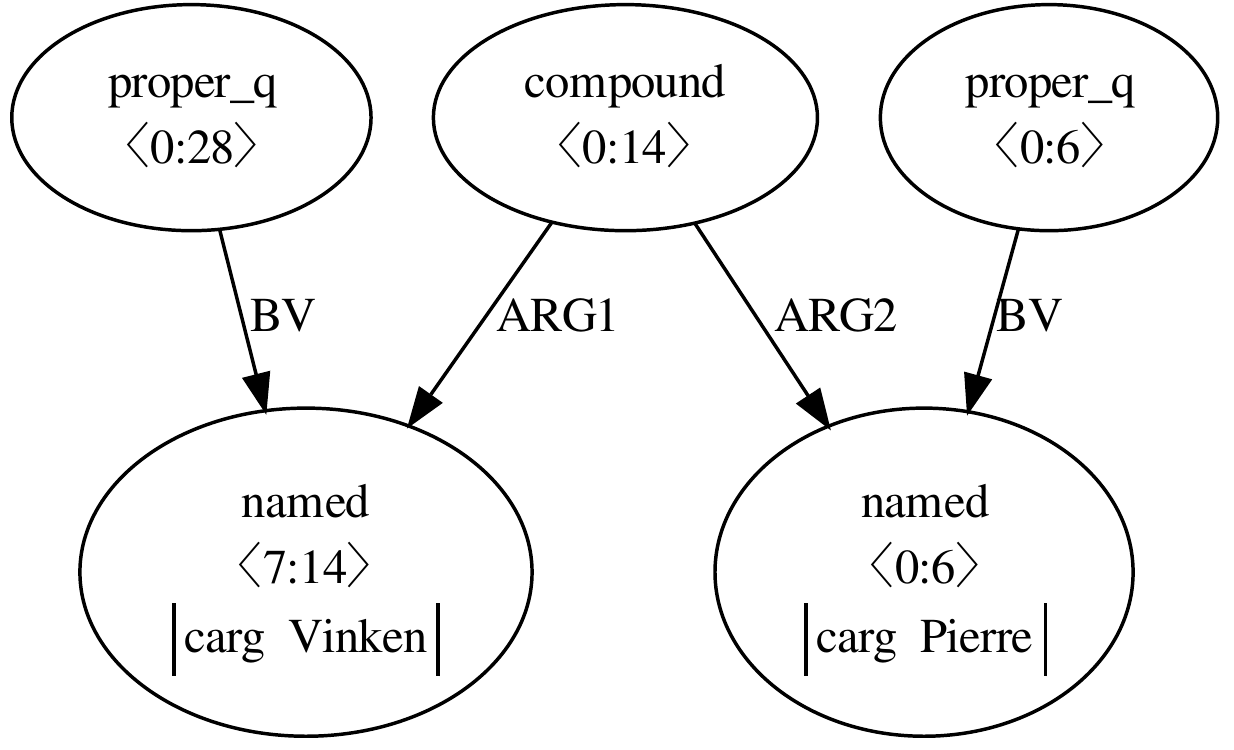}
\caption{Before reduction.}
\end{subfigure}{}
\begin{subfigure}{\linewidth}
\centering
\includegraphics[width=0.99\linewidth]{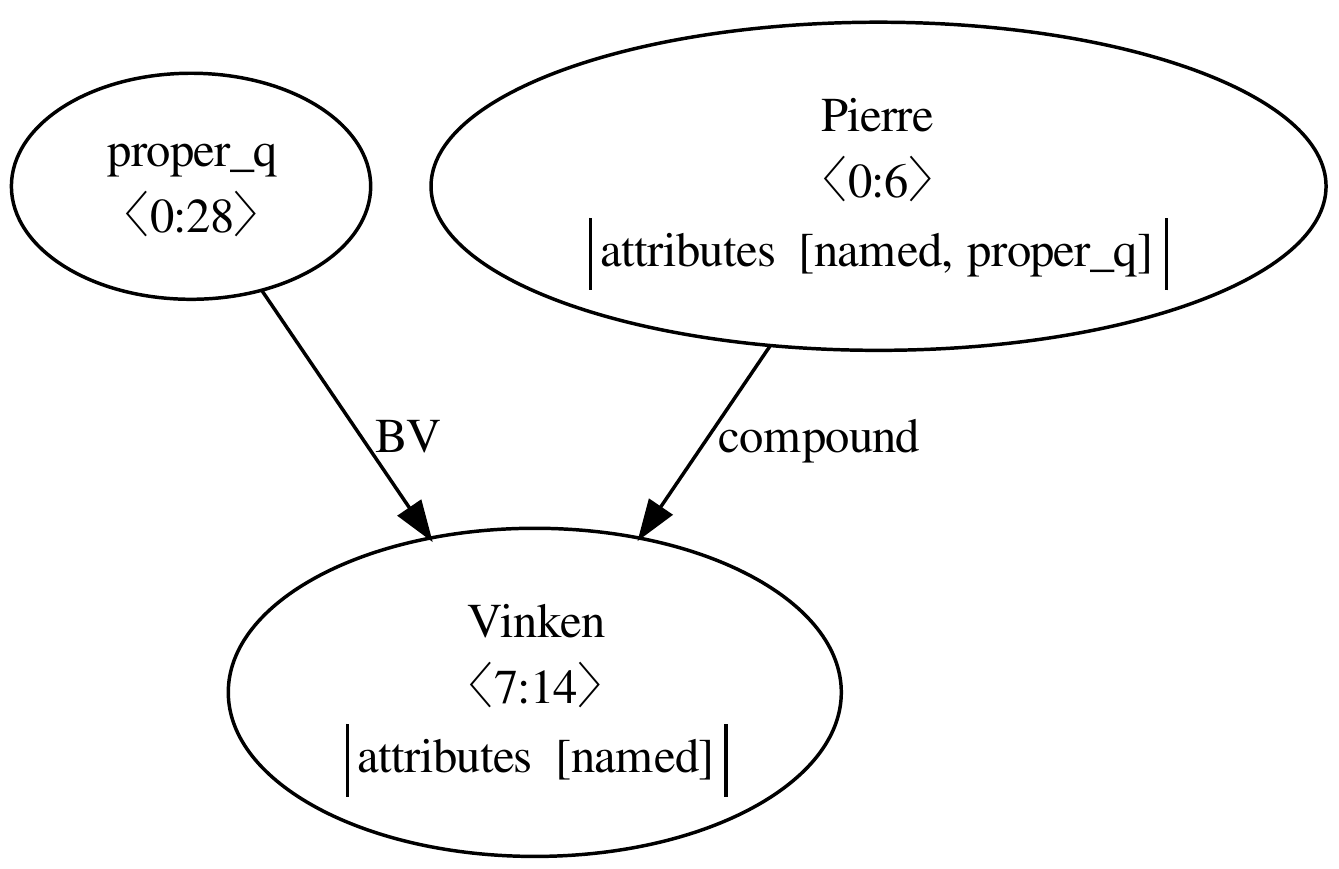}
\caption{After reduction.}
\end{subfigure}{}
\caption{An example of EDS reduction. This is a sub-graph of sentence \textit{\#20001001}.}
\label{fig:reduction}
\end{figure}

\label{sec:data_eds}
In pre-processing we first clean the companion data to make sure the tokens in the companion data is consistent with those in the MRP input. We suppose anchors are continuous for each node, so we replace the anchors with the corresponding start and end token indices. 

In EDS graphs, there are a lot of nodes without a direct mapping to individual surface tokens, which we call \textit{type} $1$ nodes. We call nodes with corresponding surface tokens \textit{type} $2$ nodes. We reduce \textit{type} $1$ nodes in two ways:
\begin{itemize}
    \item If a node $a$ of \textit{type} $1$ is connected to only one node $b$ which is of \textit{type} $2$ and has the same anchor as $a$, we reduce node $a$ into node $b$ as a special \textit{attribute} for the node.
    \item If a node $a$ of \textit{type} $1$ is connected to exactly two nodes $b$ and $c$ which are of \textit{type} 2 and have a combined anchor range that matches the anchor of $a$. We reduce node $a$ as an edge connecting $b$ and $c$ with the same label. The edge direction is decided by the labels of the edges connecting $a$ to $b$ and $c$. For example, if node $a$ has two child nodes $b$ and $c$, edge $(a,c)$ has label $ARG2$ and edge $(a,b)$ has label $ARG1$, then node $a$ will be reduced to directed edge $(b,c)$ with the label of node $a$.
\end{itemize} 
An example of the reduction is shown in Fig. \ref{fig:reduction}.
This method reduces 4 nodes on average for each graph.
We also look at nodes whose node label corresponds to a multi-word in the sentence
For example, `\_such+as' in an EDS graph corresponds to `such as' in the sentence. In such case, if the phrase has a probability over 0.5 that maps to a single node, then all words in this phrase will be combined to a single token in the sentence. 

In the post-processing, we recover reduced nodes by reversing the reduction precedure according to the node attributes and edge labels.

For UCCA, we label implicit nodes with special labels $n_i$, where $i$ is the index that the implicit node appears in the node sequence.

\section{System Description}
\begin{figure*}[t]
\centering
\includegraphics[scale=0.45]{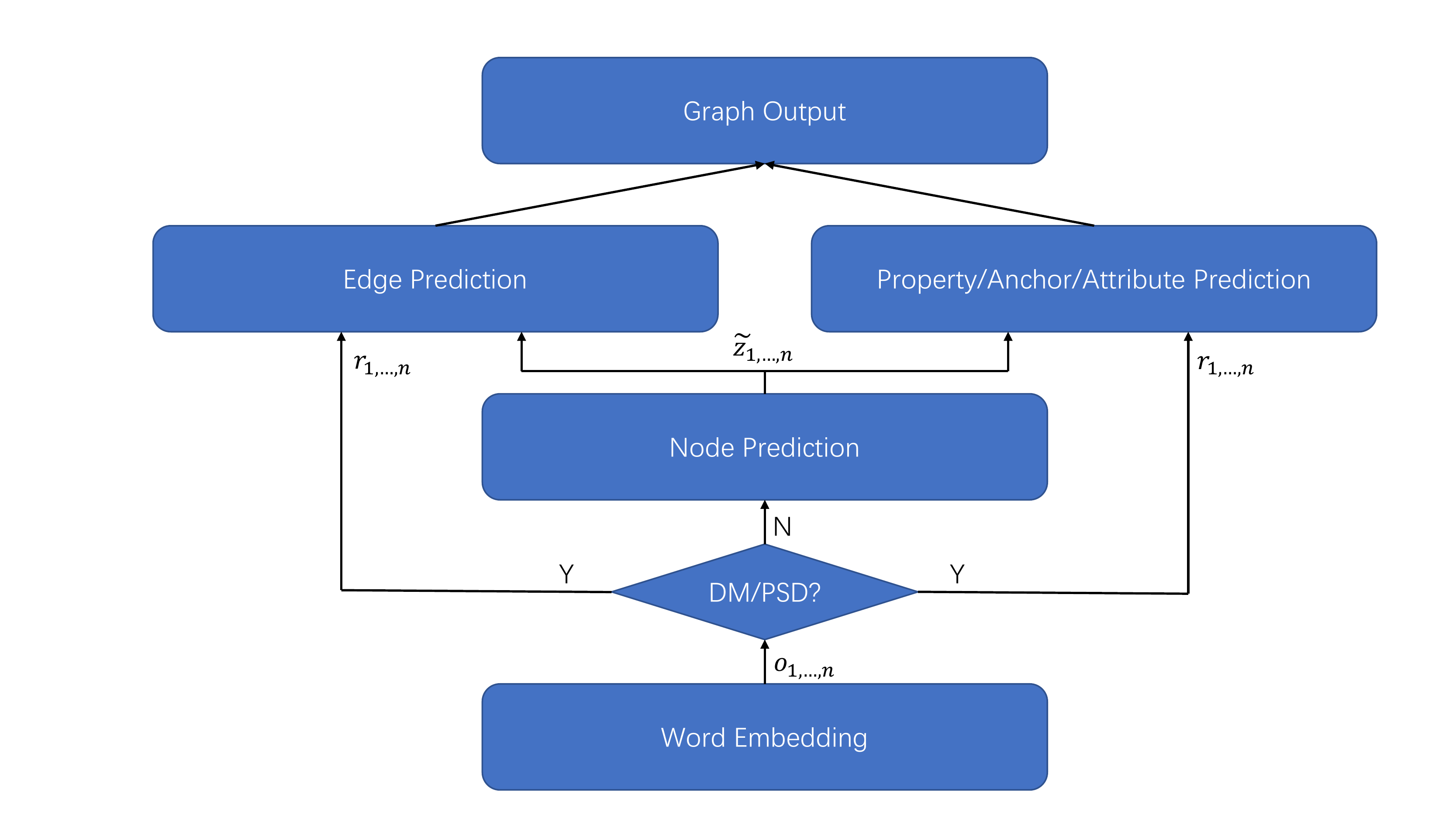}
\caption{Illustration of our system architecture.}
\label{fig:pipeline}
\end{figure*}

In this section, we describe our model for the task. We first predict the nodes of the parse graph. For DM and PSD, there is a one-to-one mapping between sentence tokens and graph nodes. For EDS, UCCA and AMR, we apply an extended pointer-generator network \cite{zhang-etal-2018-stog} for node prediction.
Given predicted nodes, we then adopt the method of second-order mean field variational inference \cite{wang-etal-2019-second} for edge prediction. Figure \ref{fig:pipeline} illustrates our system architecture.
\subsection{Word Representation}
\label{sec:embeddings}
Previous work found that various word representation could help improve parser performance. Many state-of-the-art parsers use POS tags and pre-trained GloVe \cite{pennington2014glove} embeddings as a part of the word representation. \citet{dozat2018simpler} find that character-based LSTM and lemma embeddings can further improve the performance of semantic dependency parser. \citet{zhang-etal-2018-stog} use BERT \cite{devlin-etal-2019-bert} embeddings for each token to improve the performance of AMR parsing. In our system, we find that predicted named entity tags are helpful as well. The word representation $o_i$ in our system is:
\begin{align*}
    \mathbf{o}_i=[\mathbf{o}_i^{w};\mathbf{o}_i^{\textrm{pos}};\mathbf{o}_i^{\textrm{lemmas}};\mathbf{o}_i^{pw};\mathbf{o}_i^{bw};\mathbf{o}_i^{\textrm{char}};\mathbf{o}_i^{\textrm{ne}}]
\end{align*}
where $\mathbf{o}_i^{w}$ is word embedding with random initialization, $\mathbf{o}_i^{pw}$ is pre-trained GloVe embedding and $\mathbf{o}_i^{bw}$ are BERT embedding through average pooling over subwords. $\mathbf{o}_i^{\textrm{pos}}$, $\mathbf{o}_i^{\textrm{lemmas}}$, $\mathbf{o}_i^{\textrm{char}}$, $\mathbf{o}_i^{\textrm{ne}}$ are XPOS, lemmas, character and NER embedding respectively. XPOS and lemmas are extracted from the official companion data.

\subsection{Node Prediction}
\label{sec:ptr-gen}
We use extended pointer-generator network \cite{zhang-etal-2018-stog} for nodes prediction. Given a sentence with $n$ words $\mathbf{w}=[w_1,w_2,...,w_n]$, we predict a list of nodes $\mathbf{u}=[u_1,u_2,...,u_m]$ sequentially and assign their corresponding indices $\mathbf{idx}=[idx_1,idx_2,...,idx_m]$. The indices $\mathbf{idx}$ are used to track whether a copy of a previous generated nodes or a newly generated node.
\begin{equation*}
    P(\mathbf{u})=\prod_{i=1}^mP(u_i\mid u_{<i}, idx_{<i}, \mathbf{w})
\end{equation*}
To encode the input sentence, we use a multi-layer BiLSTM fed with embeddings of the words: \begin{align}
    \label{eq:bilstm}
    R&=\mathrm{BiLSTM}(O)
\end{align}
where $O$ represents $[\mathbf{o}_1,\dots,\mathbf{o}_n]$, $\mathbf{o}_i$ is the concatenation different types of embeddings for $w_i$, and $R=[\mathbf{r}_1,\dots,\mathbf{r}_n]$ represents the output from the BiLSTM.

For the decoder, at each time step $t$, we use an $l$-layer LSTM for generating hidden states $z_t^l$ sequentially:
\begin{equation*}
    \mathbf{z}^l_t = f^l(\mathbf{z}^{l-1}_t, \mathbf{z}^l_{t-1})
\end{equation*}
where $f^l$ is the $l$-th layer of LSTM, $\mathbf{z}^l_0$ is the last hidden state $r_n$ in Eq. \ref{eq:bilstm}. $\mathbf{z}^0_t$ is the concatenation of the label embedding of node $u_{t-1}$
and attentional vector $\widetilde{\mathbf{z}}_{t-1}$. $\widetilde{\mathbf{z}}_{t}$ is defined by:

\begin{align}
    \label{eq:e_src}\mathbf{e}^t_\textrm{src} = &\mathbf{W}_{satt}^{\top}\textrm{tanh}(\mathbf{W}_\textrm{src}R + \mathbf{U}_\textrm{src}\mathbf{z}^l_t + \mathbf{b}_\textrm{src}) \\
    \label{eq:a_src}\mathbf{a}^t_\textrm{src} = &\textrm{softmax}(\mathbf{e}^t_\textrm{src}) \\
    \nonumber\mathbf{c}_t=&\sum_{i}^{n}\mathbf{a}^t_\textrm{src,i}\mathbf{r}_{i} \\
    \label{eq:z_hidden}\widetilde{\mathbf{z}}_t =& \textrm{tanh}(\mathbf{W}_c[\mathbf{c}_t;\mathbf{z}^l_t] + \mathbf{b}_c)
\end{align}
Where $\mathbf{a}^t_\textrm{src}$ is the source attention distribution, and $\mathbf{c}_t$ is contextual vector of encoder hidden layers, $\mathbf{W}_{satt}$, $\textbf{W}_{\textrm{src}}$, $\textbf{U}_{\textrm{src}}$, $\textbf{b}_{\textrm{src}}$, $\textbf{W}_{c}$, $\textbf{b}_{c}$ are learnable parameters.
The vocabulary distribution is given by:
\begin{equation}
    P_\textrm{vocab} = \textrm{softmax}(\mathbf{W}_\textrm{vocab}\widetilde{\mathbf{z}}_t + \mathbf{b}_\textrm{vocab})\label{eq:pvocab}
\end{equation}
where $\mathbf{W}_\textrm{vocab}$ and $\mathbf{b}_\textrm{vocab}$ are learnable parameters.
The target attention distribution is defined similarly as Eq. \ref{eq:e_src} and \ref{eq:a_src}:
\begin{align*}
    \mathbf{e}^t_\textrm{tgt} = &\mathbf{W}_{tatt}^{\top}\textrm{tanh}(\mathbf{W}_\textrm{tgt}\widetilde{\mathbf{z}}_{1:t-1} + \mathbf{U}_\textrm{tgt}\widetilde{\mathbf{z}}_t + \mathbf{b}_\textrm{tgt}), \\
    \mathbf{a}^t_\textrm{tgt} = &~\textrm{softmax}(\mathbf{e}^t_\textrm{tgt}),
\end{align*}
where $\mathbf{W}_{tatt}^{\top}$, $\textbf{W}_{\textrm{tgt}}$, $\textbf{U}_{\textrm{tgt}}$, $\textbf{b}_{\textrm{tgt}}$ are learnable parameters.
Finally, at each time step, we need to decide which action should be taken. Possible actions include copying an existing node from previous nodes and generating a new node whose label is either from the vocabulary or a word from the source sentence. The corresponding probability of these three actions are $p_\textrm{tgt}$, $p_\textrm{gen}$ and $p_\textrm{src}$:
\begin{equation*}
    [p_\textrm{tgt},p_\textrm{gen},p_\textrm{src}] = \textrm{softmax}(\mathbf{W}_\textrm{action}\widetilde{\mathbf{z}}_t + \mathbf{b}_\textrm{action})
\end{equation*}
where $p_\textrm{tgt}+p_\textrm{gen}+p_\textrm{src}=1$.

At time step $t$, if $u_t$ is a copy of an existing nodes, then the probability $P^\textrm{(node)}(u_t)$ and the index $idx_t$ is defined by: 
\begin{align*}
    P^\textrm{(node)}(u_t) &=p_\textrm{tgt}\sum_{i:u_i=u_t}\mathbf{a}^t_\textrm{tgt}[i]\\
    idx_t&=idx_j
\end{align*}
where $idx_j$ is the copied node index. If $u_t$ is a new node:
\begin{align*}
    P^\textrm{(node)}(u_t) &=  p_\textrm{gen}P_\textrm{vocab}(u_t) +
    p_\textrm{src}\sum_{i:w_i=u_t}\mathbf{a}^t_\textrm{src}[i]\\
    idx_t&=t
\end{align*}

\subsection{Edge Prediction}
\begin{figure*}[tb]
\centering
\includegraphics[scale=0.45]{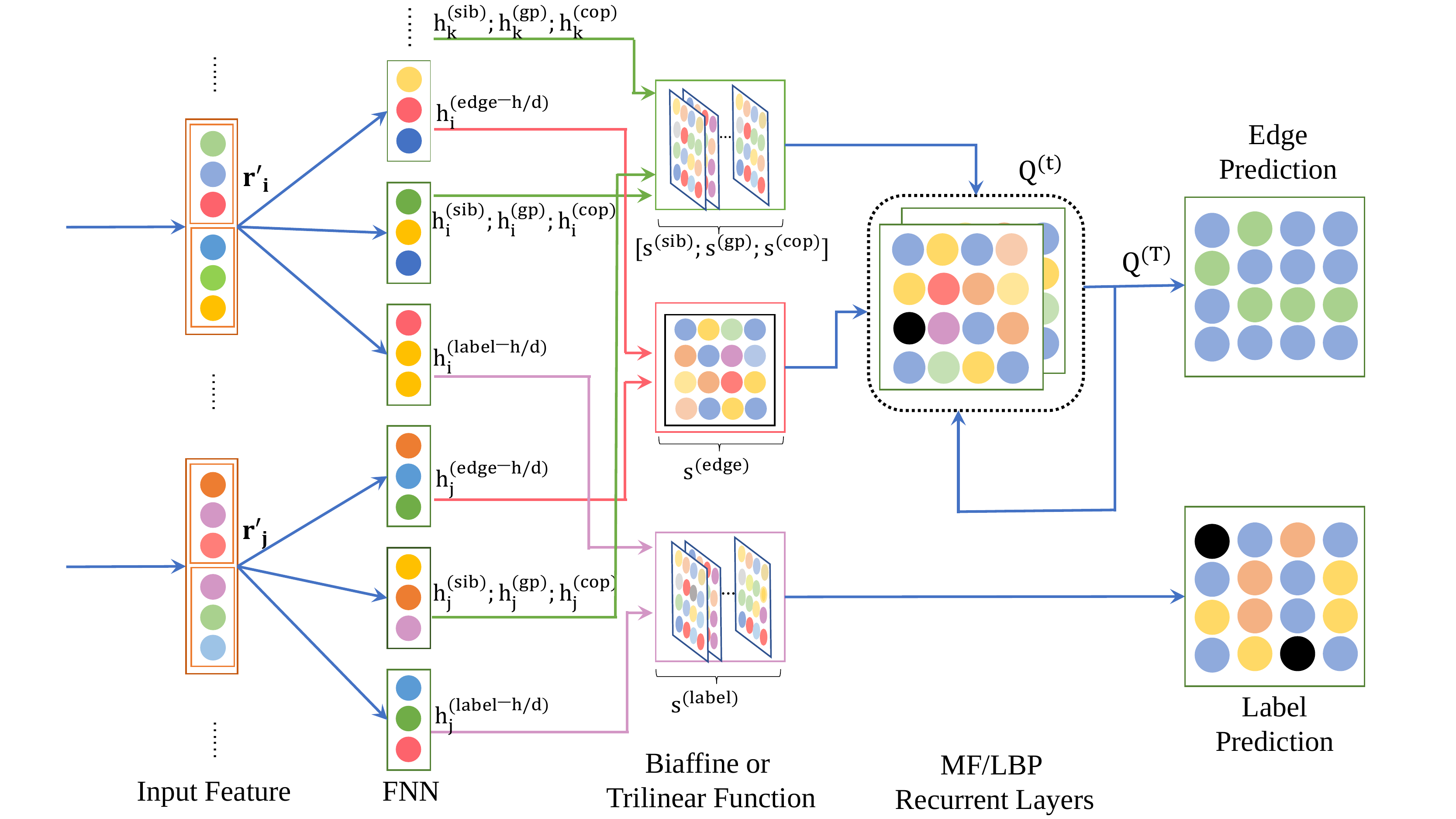}
\caption{The structure of our edge prediction module. The figure is from \citet{wang-etal-2019-second} with minor modifications.
}
\label{fig:edge_prediction}
\end{figure*}

We adopt the method presented in \citet{wang-etal-2019-second} for edge prediction, which is based on second-order scoring and inference. Suppose that we have a sequence of vector representations of the predicted nodes $[\mathbf{r}^{\prime}_1,\dots,\mathbf{r}^{\prime}_m]$, which can be the BiLSTM output $\mathbf{r}_i$ in Eq. \ref{eq:bilstm} in the cases of DM and PSD, or the extended pointer-generator network output $\widetilde{\mathbf{z}}_i$ in Eq. \ref{eq:z_hidden} in the cases of EDS, UCCA and AMR. The edge prediction module is shown in Fig. \ref{fig:edge_prediction}.

To score first-order and second-order parts (i.e., edges and edge-pairs) in both edge-prediction and label-prediction, we apply the Biaffine function \cite{dozat2016deep,dozat2018simpler} and Trilinear function \cite{wang-etal-2019-second} fed with node representations.
\begin{align}
    &\mathrm{Biaff}(\mathbf{v}_1,\mathbf{v}_2):=\mathbf{v}_1^{\top} \mathbf{U} \mathbf{v}_2 + \mathbf{b}\nonumber \\
    &\nonumber\mathbf{g}_i:=\mathbf{U}_i\mathbf{v}_i \qquad i\in[1,2,3]\\
    &\label{eq:trilin}\mathrm{Trilin}(\mathbf{v}_1,\mathbf{v}_2,\mathbf{v}_3):=\sum_{i=1}^{d}\mathbf{g}_{1i}\circ \mathbf{g}_{2i}\circ \mathbf{g}_{3i}
\end{align}
where $\mathbf{U}_i$ is a $(d\times d)$-dimensional tensor, where $d$ is hidden size and $\circ$ represents element-wise product.
We consider three types of second-order parts: siblings (sib), co-parents (cop) and grandparents (gp) \cite{martins2014priberam}. For a specific first-order and second-order \textit{part}, we use single-layer FNNs to compute a \textit{head} representation and a \textit{dependent} representation for each word, as well as a \textit{head\_dep} representation which is used for grandparent parts:
\begin{align}
    &\textrm{part}\in \{\textrm{edge},\textrm{label},\textrm{sib},\textrm{cop},\textrm{gp}\}\nonumber\\
    &\mathbf{h}_i^{\text{(part-head)}}=\text{FNN}^{\text{(part-head)}}(\mathbf{r}^{\prime}_i)\nonumber\\
    &\mathbf{h}_i^{\text{(part-dep)}}=\text{FNN}^{\text{(part-dep)}}(\mathbf{r}^{\prime}_i)\nonumber\\
    &\mathbf{h}_i^{\text{(gp-head\_dep)}}=\text{FNN}^{\text{(gp-head\_dep)}}(\mathbf{r}^{\prime}_i)\nonumber
\end{align}
We then compute the part scores as follows:
\begin{align}
    &\label{eq:edge}s_{ij}^{\textrm{(edge)}}=\mathrm{Biaff}^{\textrm{(edge)}}(\mathbf{h}_i^{\textrm{(edge-dep)}},\mathbf{h}_j^{\textrm{(edge-head)}})\\
    &\label{eq:label}\mathbf{s}_{ij}^{\textrm{(label)}}=\mathrm{Biaff}^{\textrm{(label)}}(\mathbf{h}_i^{\textrm{(label-dep)}},\mathbf{h}_j^{\textrm{(label-head)}})\\
    &\label{eq:sib}s^{(sib)}_{ij,ik} \equiv s^{(sib)}_{ik,ij}= \mathrm{Trilin}^{\text{(sib)}}(\mathbf{h}_i^{\text{\text{(head)}}},\mathbf{h}_j^{\text{\text{(dep)}}},\mathbf{h}_k^{\text{\text{(dep)}}})\\
    &s^{(cop)}_{ij,kj} \equiv s^{(cop)}_{kj,ij}=\label{eq:cop}\mathrm{Trilin}^{\text{(cop)}}(\mathbf{h}_i^{\text{(head)}},\mathbf{h}_j^{\text{\text{(dep)}}},\mathbf{h}_k^{\text{(head)}})\\
    &s^{(gp)}_{ij,jk} = \label{eq:gp}\mathrm{Trilin}^{\text{(gp)}}(\mathbf{h}_i^{\text{(head)}},\mathbf{h}_j^{\text{(head\_dep)}},\mathbf{h}_k^{\text{(dep)}})
\end{align}
In Eq. \ref{eq:edge},\ref{eq:label}, the tensor $\mathbf{U}$ in the biaffine function is $(d\times 1 \times d)$-dimensional and $(d\times c)$-dimensional, where $c$ is the number of labels. We require $j<k$ in Eq. \ref{eq:sib} and $i<k$ in Eq. \ref{eq:cop}. 


In the label-prediction module, $\mathbf{s}_{i,j}^{\text{(label)}}$ is fed into a softmax layer that outputs the probability of each label for edge $(i,j)$. In the edge-prediction module, we can view computing the edge probabilities as doing posterior inference on a Conditional Random Field (CRF).  Each Boolean variable $X_{ij}$ in the CRF indicates whether the directed edge $(i,j)$ exists. We use Eq. \ref{eq:edge} to define our unary potential $\psi_u$ representing scores of an edge and Eqs. (\ref{eq:sib}-\ref{eq:gp}) to define our binary potential $\psi_p$.
We define a unary potential $\phi_u(X_{ij})$ for each variable $X_{ij}$.
\begin{align*}
    \phi_{u}(X_{ij})=&
    \begin{cases}
    \exp(s_{ij}^{\textrm{(edge)}}) &\text{$X_{ij}=1$}\\
    1 &\text{$X_{ij}=0$}
    \end{cases}
\end{align*}
For each pair of edges $(i,j)$ and $(k,l)$ that form a second-order part of a specific $type$, we define a binary potential $\phi _{p}(X_{ij}, X_{kl})$.
\begin{align*}
\phi_{p}(X_{ij},X_{kl})&=
\begin{cases}
\exp(s^{(type)}_{ij,kl}) &\text{$X_{ij}=X_{kl}=1$}\\
1 &\text{Otherwise}
\end{cases}
\end{align*}

Exact inference on this CRF is intractable. We use mean field variational inference to approximate a true posterior distribution with a factorized variational distribution and tries to iteratively minimize their KL divergence.
We can derive the following iterative update equations of distribution $Q_{ij}(X_{ij})$ for each edge $(i,j)$.
\begin{equation}
    \begin{aligned}
    \label{eq:msg}\mathcal{F}^{(t-1)}_{ij}=&\sum_{k\neq i,j}Q^{(t-1)}_{ik}(1)s^{(sib)}_{ij,ik}+Q^{(t-1)}_{kj}(1)s^{(cop)}_{ij,kj}\\
    &+Q^{(t-1)}_{jk}(1)s^{(gp)}_{ij,jk}+Q^{(t-1)}_{ki}(1)s^{(gp)}_{ki,ij}
    \end{aligned}
\end{equation}
\begin{equation*}
\begin{aligned}
    Q_{ij}^{(t)}(0)&\propto 1\\
    Q_{ij}^{(t)}(1)&\propto \mathrm{exp} \{s^{\textrm{(edge)}}_{ij}+ \mathcal{F}^{(t-1)}_{ij}\}\
\end{aligned}
\end{equation*}
The initial distribution $Q^{(0)}_{ij}(X_{ij})$ is set by normalizing the unary potential $\phi_u(X_{ij})$. We iteratively update the distributions for $T$ steps and then output $Q^{(T)}_{ij} (X_{ij})$, where $T$ is a hyperparameter. We can then predict the parse graph by including every edge $y^{\textrm{(edge)}}_{ij}$ such that $Q^{(T)}_{ij}(1) > 0.5$.  The edge labels $y^{\textrm{(label)}}_{ij}$ are predicted by maximizing the label probabilities computed by the label-prediction module.
\begin{align*}
P(y^{\textrm{(edge)}}_{ij}|\mathbf{w})&=\textrm{softmax}(Q_{ij}^{(T)}(X_{ij}))\\
P(y^{\textrm{(label)}}_{ij}|\mathbf{w})&=\textrm{softmax}(\mathbf{s}_{ij}^{\textrm{(label)}})
\end{align*}

Note that the iterative updates in mean-field variational inference can be seen as a recurrent neural network that is parameterized by the potential functions. Therefore, the whole edge prediction module can be seen as an end-to-end neural network.

\subsection{Other Predictions}
The shared task also requires prediction of component pieces such as top nodes, node properties, node anchoring and edge attributes. In this section, we present our approaches to predicting these components.
\subsubsection*{Top Nodes}
We add an extra \textit{ROOT} node for each sentence to determine the top node through edge prediction for DM and PSD. For the other frameworks, we use the first predicted node as the top node.
\subsubsection*{Node Properties}
Node properties vary among different frameworks. For DM and PSD, we need to predict the POS and frame for each node. As DM and PSD are bi-lexical semantic graphs, we directly use the prediction of XPOS from the official companion data. We use a single layer MLP fed with word features obtained in Eq. \ref{eq:bilstm} for frame prediction. For EDS, the properties only contain `carg' and the corresponding values are related to the surface string. For example, the EDS sub-graph in Fig. \ref{fig:reduction} contains a node with label `named' which has property `carg' with a corresponding value `Pierre'. The anchor of this node matches the token `Pierre' in the sentence. We found that nodes with properties have limited types of node labels. Therefore, we exchange node labels and values for EDS nodes containing properties during training. We combine the node \textit{attributes} and value predictions described in Section \ref{sec:data_eds} together as a multi-label prediction task. We use a single layer MLP to predict node labels specially for nodes with properties. For each property value, we regard it as a node label and use the extended pointer-generator network described in Section \ref{sec:ptr-gen} to predict it. Therefore, the probability of node property prediction is:
\begin{equation}
    P_{prop} = \textrm{softmax}(\mathbf{W}_\textrm{prop}\widetilde{\mathbf{r}}^{\prime}_t + \mathbf{b}_\textrm{prop})
\end{equation}

\subsubsection*{Node Anchoring}
As DM and PSD contain only token level dependencies, we can decide a node anchor by the corresponding token. For the other frameworks, we use two biaffine functions to predict the `start token' and `end token' for each node and the final anchor range is decided by the start position of `start token' and the end position of `end token'. The biaffine function is fed by word features from the encoder RNN and node features from decoder RNN.
\begin{align}
    \nonumber&s_{ij}^{\textrm{(start/end)}}=\mathrm{Biaff}^{\textrm{(start/end)}}(\mathbf{r}_i,\widetilde{\mathbf{z}}_j)\\   
    &P_{\textrm{start/end},j} = \textrm{softmax}([s_{1j},s_{2j},\dots,s_{nj}])
    \label{eq:anchor}
\end{align}
where $i$ ranges from $1$ to $n$ and $j$ ranges from $1$ to $m$.
\subsubsection*{Edge Attributes}
Only UCCA requires prediction of edge attributes, which are the `remote' attributes of edges. We create new edge labels by combining the original edge labels and edge attributes. In this way, edge attribute prediction is done by edge label prediction.

\subsection{Learning}
Given a gold graph $y^{\star}$, we use the cross entropy loss as learning objective:
\begin{align*}
\mathcal{L}^{\textrm{(edge)}} (\theta) &= -\sum_{i,j} \log(P_\theta (y_{ij}^{\star\textrm{(edge)}}|\mathbf{w}))\\
\mathcal{L}^{\textrm{(label)}} (\theta) &= -\sum_{i,j}\mathbbm{1}(y_{ij}^{\star\textrm{(edge)}}) \log(P_\theta (y_{ij}^{\star\textrm{(label)}}|\mathbf{w}))
\end{align*}
\begin{align*}
\mathcal{L}^{\textrm{(prop)}} (\theta) &= -\sum_{i,k} \log(P_\theta (y_{ik}^{\star\textrm{(prop)}}|\mathbf{w}))\\
\mathcal{L}^{\textrm{(anchor)}} (\theta) &= -\sum_{i}\sum_{j\in \{\textrm{start},\textrm{end}\}} (\log(P_\theta (y_{i}^{\star\textrm{(j)}}|\mathbf{w}))
\end{align*}
where $\theta$ is all the parameters of the model, $\mathbbm{1}(\mathcal{X})$ is an indicator function of whether $\mathcal{X}$ exists in the graph, $i,j$ range over all the nodes and $k$ ranges over all possible \textit{attributes} in the graph.
The total loss is defined by:
\begin{align*}
\mathcal{L} = &\lambda_1\mathcal{L}^{\textrm{(edge)}} + \lambda_2\mathcal{L}^{\textrm{(label)}} +\mathbbm{1}(y^{\star\textrm{(prop)}})\lambda_3\mathcal{L}^{\textrm{(prop)}} \\
&+ \mathbbm{1}(y^{\star\textrm{(anchor)}})\lambda_4\mathcal{L}^{\textrm{(anchor)}}
\end{align*}
where $\lambda_{1,\dots,4}$ are hyperparameters. For DM and PSD, we tuned on $\lambda_1$, $\lambda_2$ and $\lambda_3$. For other frameworks, we set all of them to be $1$.

\section{Experiments and Results}

\begin{table}[t!]
\centering
\small
\begin{tabular}{lccccc}
\hline \hline
 & DM & PSD & EDS & UCCA & AMR\\
 \hline
Ours-all & 94.88 & 89.49 & 86.90 & - & 63.59\\
Best-all & \textbf{95.50} & \textbf{91.28} & \textbf{94.47} & \textbf{81.67} & \textbf{73.38}\\
\hline
Ours-lpps & 94.28 & 85.22 & 87.49 & - & 66.82 \\
Best-lpps & \textbf{94.96} & \textbf{88.46} & \textbf{92.82} & \textbf{82.61} & \textbf{73.11}\\
\hline\hline
\end{tabular}
\caption{Comparison of cross-framework F1 scores achieved by our system and best scores of other teams for each metric. \textit{all} represents the F1 score over the full test set for each framework. \textit{lpps} represents a 100-sentence sample from the little prince containing graphs over all the frameworks.}
\label{tab:main}
\end{table}

\subsection{Training}
For DM, PSD and EDS, we used the same dataset split as previous approaches \cite{martins2014priberam, du2015peking} with 33,964 sentence in the training set and 1,692 sentences in the development set. For each of the other frameworks, we randomly chose 5\% to 10\% of the training set as the development set. We additionally removed graphs with more than 60 nodes (or with input sentences longer than 60 words for DM and PSD). We trained our model for each framework separately and used Adam \cite{kingma2014adam} to optimize our system, annealing the learning rate by 0.5 for 10,000 steps. We trained the model for 100,000 iterations with a batch size of 6,000 tokens and terminated with 10,000 iterations without improvement on the development set.

\begin{table*}[ht!]
\centering
\begin{tabular}{lcccccc}
\hline \hline
 & tops & labels & properties & anchors & edges & average\\
 \hline
Ours-all & \textbf{93.68} & 90.51 & \textbf{95.16} & 98.38 & \textbf{92.32} & 94.32\\
Best-all & 93.23 & \textbf{96.34} & 94.93 & \textbf{98.74} & 92.08 & \textbf{94.76}\\
\hline
Ours-lpps & \textbf{99.00} & 87.26 & \textbf{94.53} & \textbf{99.36} & \textbf{93.92} & 94.03 \\
Best-lpps & 96.48 & \textbf{94.82} & 94.36 & 99.04 & 93.28 & \textbf{94.64}\\
\hline\hline
\end{tabular}
\caption{Comparison of cross-framework F1 scores achieved by our system and best scores of the other teams for each evaluation component on DM. \textit{average} is the micro-average among all components.}
\label{tab:component_dm}
\end{table*}

\begin{table*}[ht!]
\centering
\begin{tabular}{lcccccc}
\hline \hline
 & tops & labels & properties & anchors & edges & average\\
 \hline
Ours-all & 95.68 &	84.79 & 91.83 & 97.66 & \textbf{79.50} & 88.77\\
Best-all & \textbf{95.83} &	\textbf{94.68} &	\textbf{92.38} &	\textbf{98.35} &	79.44 & \textbf{90.76}\\
\hline
Ours-lpps & 96.00&	76.72&	84.73&	97.61&	\textbf{79.80} & 85.22\\
Best-lpps & \textbf{96.40}&	\textbf{92.04}&	\textbf{86.00}&	\textbf{98.46}&	79.18 & \textbf{88.40}\\
\hline\hline
\end{tabular}
\caption{Comparison of cross-framework F1 scores achieved by our system and best scores of the other teams for each evaluation component on PSD.}
\label{tab:component_psd}
\end{table*}

\begin{table}[ht!]
\small
\centering
\begin{tabular}{lcccccc}
\hline \hline
 & \multicolumn{2}{c}{DM} & \multicolumn{2}{c}{PSD} & \multicolumn{2}{c}{Avg}\\
 & all & lpps & all & lpps & all & lpps\\
 \hline
Ours & \textbf{92.98} & \textbf{94.46} & 81.61 & \textbf{81.91}& \textbf{87.30} & \textbf{88.19}\\
Best & 92.52 & 93.68 & \textbf{81.66} & 81.47 & 87.09 & 87.58\\
\hline\hline
\end{tabular}
\caption{Comparison of in-framework labeled F1 scores by our system and best scores over the other teams. Note that the \textit{Best} scores are not only from a single system.}
\label{tab:in-framework}
\end{table}

\subsection{Main Results}
\label{sec:main}
Due to an unexpected bug in UCCA anchor prediction, we failed to submit our UCCA prediction. Our results are still competitive to those of the other teams and we get the \nth{3} place for the DM framework in the official metrics. The main result is shown in Table \ref{tab:main}.
Our system performs well on the DM framework with an F1 score only 0.4 percent F1 below the best score on DM. Note that our system does not learn to predict node labels for DM and PSD and simply uses lemmas from the companion data as node labels. We find that compared to gold lemmas from the original SDP dataset, lemmas from the companion data have only 71.4\% accuracy. We believe that it is the main reason for the F1 score gap between our system and the best one on DM and PSD. A detailed comparison between each component will be discussed in Section \ref{sec:detail}. For PSD, EDS and AMR graph, our system ranks \nth{6}, \nth{5} and \nth{7} among 13 teams.

\subsection{Analysis}
\subsubsection*{DM and PSD}
\label{sec:detail}
Table \ref{tab:component_dm} and \ref{tab:component_psd} show detailed comparison for each evaluation component for DM and PSD. For DM, our system outperforms systems of the other teams on tops, properties and edges prediction and is competitive on anchors. For PSD, our system is also competitive on all the components except labels. There is a large gap in the performance of node label prediction between our system and the best one on both DM and PSD, we believe adding an MLP layer for label prediction would diminish this gap.

Table \ref{tab:in-framework} shows the performance comparison on in-framework metrics for DM and PSD. For DM, our system outperforms the best of the other systems by 0.5 and 0.8 F1 scores on \textit{all} and \textit{lpps} test sets. For PSD, our system outperforms the best of the other systems by 0.4 F1 score for \textit{lpps} and only 0.05 F1 score below the best score for \textit{all}. 

\subsubsection*{AMR}
\label{sec:amr}
\begin{table}[t!]
\centering
\begin{tabular}{lc}
\hline \hline
Model & Smatch\\
 \hline
\citet{zhang-etal-2018-stog} & 69.1\\
Ours & 69.3\\
\hline\hline
\end{tabular}
\caption{Smatch F1 score on AMR development set. We compare the results without post-processing.}
\label{tab:amr-smatch}
\end{table}

\begin{table}[t!]
\centering
\begin{tabular}{lcc}
\hline \hline
Set & MRP & Smatch\\
 \hline
test & 63.59 & 63.08\\
dev & 72.03 & 71.55 \\
\hline\hline
\end{tabular}
\caption{MRP and Smatch score on the development set and the test set.}
\label{tab:amr-res}
\end{table}

For AMR graph prediction, our node prediction module is based on \citet{zhang-etal-2018-stog}, but our edge prediction module is based on the second-order method of \citet{wang-etal-2019-second}. To verify the effectiveness of second-order edge prediction, we compare the performances on the development set of our model and \citet{zhang-etal-2018-stog}. The result is shown in Table \ref{tab:amr-smatch}. The result shows that our second-order edge prediction is useful not only on the SDP frameworks but also on the AMR framework. 

From the official results on the test sets, we find it surprising that there is a huge gap between the test and development results on both the MRP and the Smatch \cite{cai-knight-2013-smatch} scores, as shown in Table \ref{tab:amr-res}. In future work, we will figure out the reason behind this problem.

\subsubsection*{EDS}
For EDS, our parser ranks \nth{5}. There are multiple details of our parser that can be improved. For example, our anchor prediction module described in Eq. \ref{eq:anchor} (ranking \nth{4} in the task) may occasionally predict an end anchor positioned before a start anchor, which would be rejected by the evaluation system. This can be fixed by adding constraints.

\subsubsection*{UCCA}
For UCCA, we failed to submit the result because of the same reversed start-end anchor predictions, which prevents us from obtaining an MRP score. 

\subsection{Ablation Study}

\begin{table}[t!]
\centering
\begin{tabular}{lc}
\hline \hline
 & LF1\\
 \hline
Baseline & 93.41\\
Base-fixed & 94.17\\
Base-tuned & 94.22\\
Base-fixed + Glove & 94.45\\
Base-tuned + Glove & 94.48\\
Large-fixed + Glove & 94.62\\
Large-tuned + Glove & 94.64\\
Large-fixed + Glove + Lemma & 95.10\\
Large-fixed + Glove + Lemma + Char & 95.22\\
\hline
ELMo + Large-fixed + Glove + Lemma & 94.78\\
ELMo + Glove + Lemma + Char & 95.06\\
\hline
BERT-First & 95.22\\
BERT-Avg & 95.28\\
BERT-Avg + dep-tree & 95.30\\
\hline\hline
\end{tabular}
\caption{Comparing Labeled F1 scores of models with different types of embedding combinations on the development set of the gold DM dataset. \textit{Baseline} represents the parser of \citet{wang-etal-2019-second}. \textit{Base} represents the pre-trained BERT-Base uncased model and \textit{Large} represents the pre-trained BERT-Large uncased model. \textit{fixed} and \textit{tuned} represents whether to fine-tune the BERT model. \textit{BERT} in the last block represents the last embedding combination (Large-fixed + Glove + Lemma + Char) in the first block. \textit{First} represents first subtoken pooling, \textit{Avg} represents average pooling over subtokens. \textit{dep-tree} represents adding dependency information into embeddings. For each case, we report the highest Labeled F1 score on the development set in our experiments.}
\label{tab:bert}
\end{table}

\subsubsection*{BERT with Other Embeddings}
We use BERT \cite{devlin-etal-2019-bert} embedding in our model. We compared the performance of DM in the original SDP dataset with different subtoken pooling methods, and we also explored whether combining other embeddings such as pre-trained word embedding Glove \cite{pennington2014glove} and contextual embedding ELMo \cite{peters-etal-2018-deep} will further improve the performance. The detailed results are shown in table \ref{tab:bert}. We found that Glove, lemma and character embeddings are helpful for DM and fine-tuning on the training set slightly improves the performance. ELMo embedding is also helpful but cannot outperform BERT embedding. However, the performance dropped when ELMo embedding and BERT embedding are combined. We speculate that the drop is caused by the conflict between the two types of contextual information. For subtoken pooling, we compared the performance of using first subtoken pooling and average pooling as token embedding. We found that average pooling is slightly better than first pooling. For syntactic information, we encode each head word and dependency label as embeddings and concatenate them together with other embeddings. The result shows that syntactic information as embeddings is not very helpful for the task. We will try other methods utilizing syntactic information in future work.

\begin{table}[t!]
\centering
\begin{tabular}{lcc}
\hline \hline
 & DM & PSD\\
 \hline
basic & 96.01& 90.80 \\
+lemma & 96.09 & 90.79\\
+ner & 96.07 & 90.80\\
+lemma \& ner & \textbf{96.16}& \textbf{90.88}\\
\hline\hline
\end{tabular}
\caption{F1 score averaged over the labeled F1 score and the frame F1 score on the development sets of DM and PSD. \textit{basic} represents our model with embeddings described in \ref{sec:embeddings} except lemma and named entity embeddings.}
\label{tab:ner-ab}
\end{table}

\subsubsection*{Lemma and Named Entity Tags}
\citet{dozat2018simpler} found that gold lemma embedding is helpful for semantic dependency parsing. However, in section \ref{sec:main}, we note that the lemmas from the official companion data have only 71.4\% accuracy compared to lemmas in gold SDP data, which makes lemma embeddings less helpful for parsing. We found that one of the difference is about the lemma annotations of entities, for example, lemmas of ``Pierre Vinken'' are ``Pierre'' and ``Vinken'' in the companion data while the lemmas are named-entity-like tags ``Pierre'' and ``\_generic\_proper\_ne'' in the original SDP dataset. Based on this discovery, we experimented on the influence of named entity tags on parsing performance. We used Illinois Named Entity Tagger \cite{RatinovRo09} in white list to predict named entity tags and compared the performance on the development sets of DM and PSD. The result is shown in table \ref{tab:ner-ab}. We tuned the hyperparameters for all the embedding conditions in the table, and we found that adding lemma or named entity embeddings results in a slight improvement on DM but does not help on PSD. With both lemma and named entity embeddings, there is a further improvement on both DM and PSD, which shows the named entity tags are helpful for semantic dependency parsing. As a result, we apply named entity information in parsing other frameworks.

\section{Conclusion}
In this paper, we present our graph-based parsing system for MRP 2019, which combines two state-of-the-art methods for sequence to graph node generation and second-order edge inference. The result shows that our system performs well on the DM and PSD frameworks and achieves the best scores on the in-framework metrics. For future work, we will improve our system to achieve better performance on all these frameworks and explore cross-framework multi-task learning. Our code for DM and PSD is available at \url{https://github.com/wangxinyu0922/Second_Order_SDP}.

\bibliography{sdp,ltg,conll,mrp}
\bibliographystyle{acl_natbib}

\end{document}